\documentclass[10pt,twocolumn,letterpaper]{article}

\usepackage{cvpr}      
\usepackage[accsupp]{axessibility}
 
\usepackage{graphicx}
\usepackage{amsmath}
\usepackage{amssymb}
\usepackage{booktabs}
\usepackage{makecell}
\usepackage{multirow}
\usepackage{bm}

\definecolor{cvprblue}{rgb}{0.21,0.49,0.74}
\usepackage[pagebackref,breaklinks,colorlinks,allcolors=cvprblue]{hyperref}

\def\paperID{13648} 
\def\confName{CVPR}
\def\confYear{2025}

\title{PhysVLM: Enabling Visual Language Models to Understand Robotic Physical Reachability}

\author{
Weijie Zhou\textsuperscript{1,2}, Manli Tao\textsuperscript{2}, Chaoyang Zhao\textsuperscript{2,3,†}, Haiyun Guo\textsuperscript{2}, \\
Honghui Dong\textsuperscript{1}, Ming Tang\textsuperscript{2}, Jinqiao Wang\textsuperscript{1,2,3,4,†} \\
\textsuperscript{1} School of Traffic and Transportation, Beijing Jiaotong University \\ 
\textsuperscript{2} Foundation Model Research Center, Institute of Automation, Chinese Academy of Sciences \\
\textsuperscript{3} ObjectEye Inc. \quad
\textsuperscript{4} Guangdong Provincial Key Laboratory of Intellectual Property \& Big Data, \\
\hspace*{1.2em}Guangdong Polytechnic Normal University \\
{\tt\small †Corresponding authors: chaoyang.zhao@nlpr.ia.ac.cn, jqwang@nlpr.ia.ac.cn}
}

\begin{document}

\maketitle

\begin{abstract}
Understanding the environment and a robot's physical reachability is crucial for task execution. While state-of-the-art vision-language models (VLMs) excel in environmental perception, they often generate inaccurate or impractical responses in embodied visual reasoning tasks due to a lack of understanding of robotic physical reachability. To address this issue, we propose a unified representation of physical reachability across diverse robots, i.e., Space-Physical Reachability Map (S-P Map), and PhysVLM, a vision-language model that integrates this reachability information into visual reasoning. Specifically, the S-P Map abstracts a robot's physical reachability into a generalized spatial representation, independent of specific robot configurations, allowing the model to focus on reachability features rather than robot-specific parameters. Subsequently, PhysVLM extends traditional VLM architectures by incorporating an additional feature encoder to process the S-P Map, enabling the model to reason about physical reachability without compromising its general vision-language capabilities. To train and evaluate PhysVLM, we constructed a large-scale multi-robot dataset, Phys100K, and a challenging benchmark, EQA-phys, which includes tasks for six different robots in both simulated and real-world environments. Experimental results demonstrate that PhysVLM outperforms existing models, achieving a 14\% improvement over GPT-4o on EQA-phys and surpassing advanced embodied VLMs such as RoboMamba and SpatialVLM on the RoboVQA-val and OpenEQA benchmarks. Additionally, the S-P Map shows strong compatibility with various VLMs, and its integration into GPT-4o-mini yields a 7.1\% performance improvement.

\end{abstract}

\section{Introduction}
\label{sec:intro}

\begin{figure}[ht] 
\vskip 0.2in
\begin{center}
\centerline{\includegraphics[width=1\linewidth]{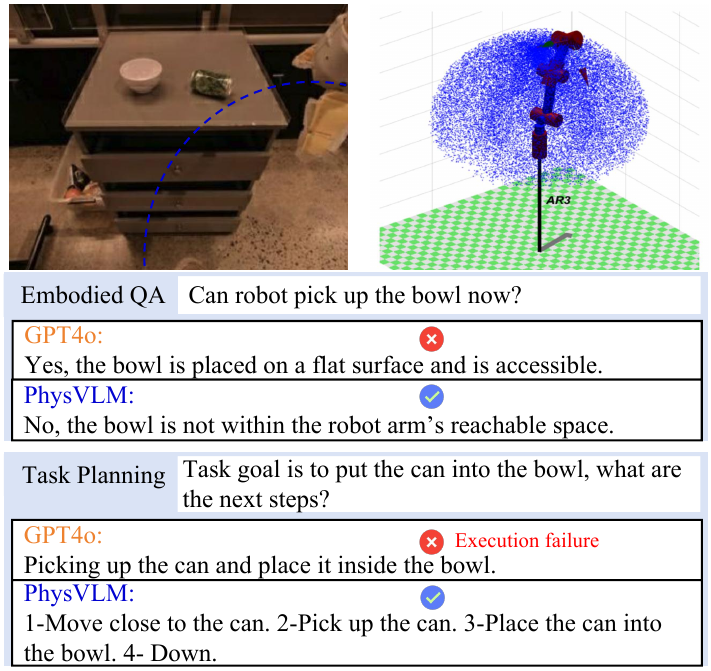}}
\caption{Existing VLM models, such as GPT-4o, may generate inaccurate or impractical responses due to poor comprehension of robotic physical reachability. The proposed PhysVLM integrates vision-language capabilities with an understanding of robotic physical reachability.}
\label{image1}
\end{center}
\vskip -0.2in
\end{figure}

Accurate perception of physical reachability is essential for robots to perform tasks effectively. Similar to how humans adjust their actions based on bodily conditions and environmental factors, robots must account for their physical reachability within an environment to ensure efficient and reliable task execution. For instance, in grasping tasks, a robot that fails to assess its reachability may attempt to grasp an object from an unreachable position, leading to task failure or equipment damage \cite{physical1, physical2}. Thus, enhancing a robot's understanding of physical reachability is crucial for successful task planning and execution in complex environments \cite{ws1, workspace_structure}.

Vision-Language Models (VLMs) have shown remarkable progress in environmental understanding \cite{lisa, llava, llavanext, internvl, qwen2vl}, and many studies have applied these models to embodied AI to assist robots in perceiving environments and planning tasks \cite{elmm1, elmm2, elmm3, elmm4, openeqa}. However, while VLMs excel in general environmental perception, they often struggle with tasks that require an understanding of robotic physical reachability (see Figure \ref{image1}). We identify two key challenges that must be addressed for VLMs to be effective in robotic tasks: (1) \textbf{how to develop a unified and efficient representation of physical reachability.} Robots vary significantly in size, joint types, and other characteristics, making it difficult for VLMs to directly learn these differences; (2) \textbf{how to enable VLMs to improve their understanding of physical reachability without compromising general vision-language capabilities.} Existing VLMs typically combine pre-trained unimodal encoders for vision and language tasks. However, introducing a new modality like physical reachability requires careful architectural and training adjustments to ensure the model can reason about reachability while maintaining its general capabilities.

To address these challenges, we propose the Space-Physical Reachability Map (S-P Map), a unified representation that abstracts the physical reachability of diverse robots into a generalized spatial form. The S-P Map is generated by combining robot parameters with egocentric depth images, but crucially, the model learns to focus on the abstracted reachability features (i.e., the gray regions in the S-P Map) rather than the specific robot configurations. This abstraction allows the model to generalize across different robots, as it only needs to reason about which areas are reachable, independent of the robot's specific characteristics. We introduce PhysVLM, a vision-language model that extends traditional VLM architectures by incorporating an additional feature encoder to process the S-P Map. This design enables PhysVLM to integrate physical reachability information into its reasoning process without compromising its general vision-language capabilities. To train and evaluate PhysVLM, we constructed a large-scale multi-robot dataset, Phys100K, and a challenging benchmark, EQA-phys, which includes tasks for six different robots in both simulated and real-world environments.

We summarize our contributions as follows:
\begin{itemize}
\item We propose a unified and robot-agnostic formulation, the Space-Physical Reachability Map (S-P Map), which abstracts robotic physical reachability in a way that is independent of specific robot configurations, promoting the learning of generalized features.
\item We introduce PhysVLM, a vision-language model that integrates physical reachability with general vision-language capabilities via an additional feature encoder, improving task execution reliability.
\item We release the EQA-phys benchmark, which includes six robots and 1.3K question-answer pairs, designed to test the model's understanding of physical reachability in simulated and real-world environments.
\item Our model achieves a 14\% improvement over GPT-4o on the EQA-phys benchmark. In embodied visual reasoning tasks on the RoboVQA-val and OpenEQA benchmarks, it outperforms advanced embodied VLMs such as RoboMamba and SpatialVLM. Furthermore, the S-P Map demonstrates strong compatibility with various VLMs, and integrating it into GPT-4o-mini results in a 7.1\% performance improvement.
\end{itemize}

\section{Related Work}

\subsection{VLMs in Robotics}

Embodied Question Answering (EQA) tasks require agents to interact with environments to answer questions \cite{palme, eqa1, eqa2}. RoboVQA offers a large, diverse dataset for robotic visual question answering. 3D-VLA \cite{3dvla} integrates 3D perception with a generative world model for embodied reasoning, while SpatialVLM \cite{spatialvlm} enhances VLMs' spatial understanding using extensive 3D data.

Robot task planning involves sequencing subtasks to achieve goals \cite{plan1, plan2, plan3}. Code as Policies (CaP) \cite{cap} employs OpenAI's Codex (code-davinci-002) to generate planning code. SayCan \cite{saycan} combines the Pathways Language Model (PaLM) \cite{palm} with robotic affordances to create feasible action plans based on the robot's capabilities. However, these methods often assume all objects are within the robot's operational area, ignoring physical reachability and potentially leading to suboptimal or infeasible plans.

\subsection{Understanding Physical Reachability}

Recent studies use voxel grids with open-vocabulary detection models to assign task-specific attributes, enabling environmental constraint understanding. ReKep \cite{rekep} generates keypoint proposals and constraints using voxel grids and VLMs, while VoxPoser \cite{voxposer} synthesizes robot trajectories by integrating OWL-ViT \cite{owlvit} and VLMs with voxel-based environment representations. However, these approaches focus on environment modeling without explicitly addressing the robot’s physical reachability.

Explicitly representing reachable workspaces remains challenging. Reachability maps \cite{workspace_structure} model spatial capabilities, and occupancy grids \cite{whole_body_reach} account for obstacles to ensure safe navigation. Additionally, methods like online model predictive control with offline workspace analysis \cite{ws1} and Reachability Expression-based Motion Planning (REMP) \cite{ws2} address workspace constraints. Despite these advancements, integrating physical reachability into visual reasoning for complex embodied tasks is still limited, primarily due to the lack of large-scale datasets that include robotic physical parameters in VLM pretraining.

\begin{figure*}[ht]
\vskip 0.2in
\begin{center}
\centerline{\includegraphics[width=1\linewidth]{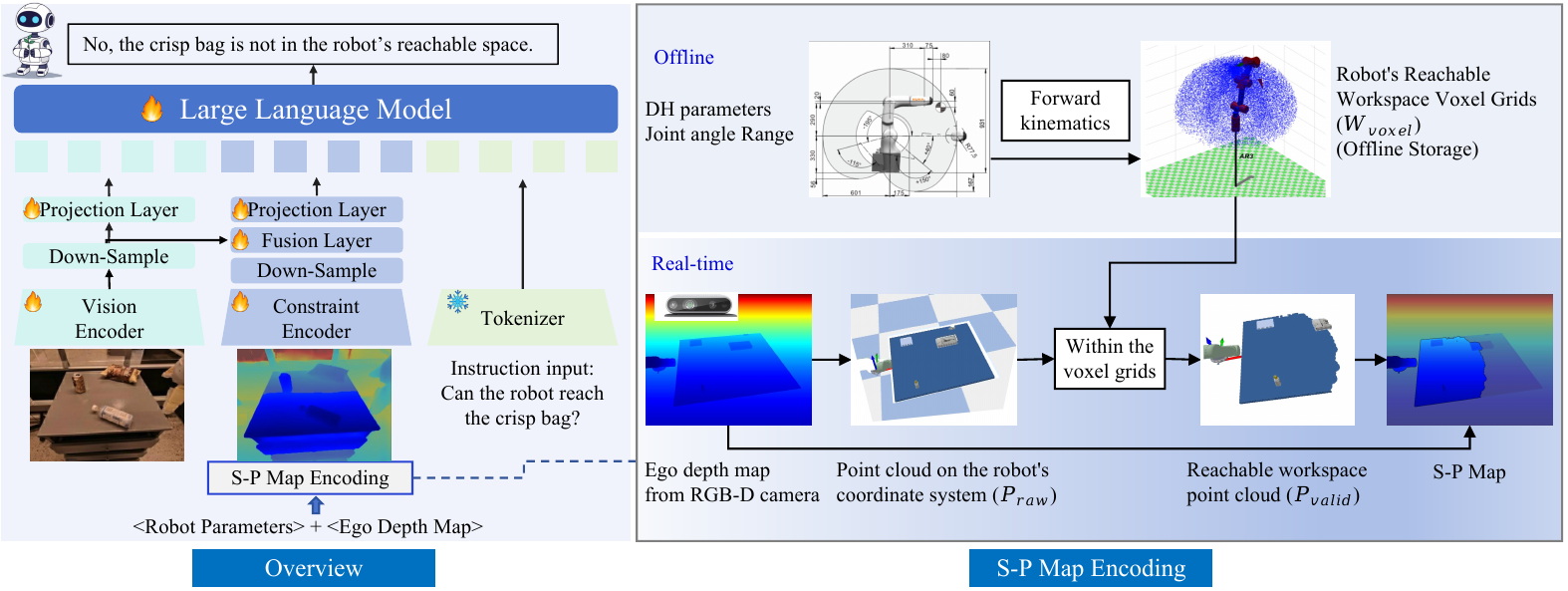}}
\caption{Overview of PhysVLM. Starting with robot parameters and an egocentric depth map, an S-P Map is generated via unified physical reachability encoding. Using the S-P Map, image, and instruction text, PhysVLM generates textual output considering the robot's physical reachability.}
\label{image2}
\end{center}
\vskip -0.2in
\end{figure*}

\section{Method}
\label{3}

PhysVLM is a large-scale vision-language model designed for visual reasoning that accounts for physical constraints in embodied tasks. As illustrated in Figure \ref{image2}, PhysVLM integrates instruction text, visual input (RGB image), and an S-P Map abstracts the robotic physical reachability into a unified spatial representation. By combining these inputs, PhysVLM generates responses consistent with both the visual context and the robot's physical reachability, without being tied to specific robot configurations. The S-P Map is constructed using a unified physical reachability encoding method, which abstracts the physical parameters of various robots alongside their egocentric depth maps into a generalized form. This abstraction allows the model to generalize across different robots, addressing the challenge of learning and reasoning about physical reachability in a robot-agnostic manner.

In this section, we present the core components of PhysVLM. Section~\ref{3.1} describes the S-P Map encoding method, Section~\ref{3.2} describes the model architecture, and Section~\ref{3.3} discusses the training procedures.

\subsection{S-P Map Encoding}
\label{3.1}

As illustrated in Figure \ref{image2}, we model the physical reachability of various robots using a unified approach that abstracts robot-specific parameters into a generalized spatial representation. This abstraction allows the model to focus on the spatial regions that are physically reachable, independent of the specific robot configuration.

\begin{equation}
\text{S-P Map} = F\left( \mathcal{P}_{\text{raw}}, \{\theta_i^{\text{min}}, \theta_i^{\text{max}}\}, \text{DH}, \mathbf{E} \right),
\end{equation}

where $\mathcal{P}_{\text{raw}}$ denotes the raw point cloud data from the robot's RGB-D camera. $\{\theta_i^{\text{min}}, \theta_i^{\text{max}}\}$ represents the range of motion for each joint $i$. $\text{DH}$ refers to the Denavit-Hartenberg parameters that describe the geometric structure of each joint, and $\mathbf{E}$ is the extrinsic calibration matrix that transforms coordinates from the camera to the robot's coordinate system. The function $F$ maps these inputs to produce the S-P Map, which abstracts the robot's physical reachability into a spatial form that is independent of the specific robot configuration.

Consider a robot arm with \(n\) degrees of freedom, where each joint \(i\) has DH parameters \(\{\theta_i, d_i, a_i, \alpha_i\}\) (\(\theta_i\) is the joint angle, \(d_i\) is the offset along the z-axis, \(a_i\) is the link length, and \(\alpha_i\) is the twist angle). The homogeneous transformation matrix for each joint is defined as:

\begin{equation}
\mathbf{T}_i = G(\theta_i, d_i, a_i, \alpha_i),
\end{equation}

where \( G \) is the standard Denavit-Hartenberg transformation function. By multiplying the transformation matrices of all joints, we obtain the transformation matrix from the base frame to the end-effector frame:

\begin{equation}
\mathbf{T} = \mathbf{T}_1 \mathbf{T}_2 \dots \mathbf{T}_n.
\end{equation}

To generate joint configurations, we sample the joint angles \(\theta_i\) from their respective motion ranges \(\theta_i \in [\theta_i^{\text{min}}, \theta_i^{\text{max}}]\), resulting in configurations \(\{\theta_1, \theta_2, \dots, \theta_n\}\). By substituting these joint configurations into the forward kinematics equations, we compute the corresponding end-effector positions:

\begin{equation}
\mathcal{W}_{\text{voxel}} = \left\{ \mathbf{p} \ \bigg| \ \mathbf{p} = \mathbf{T}(\theta_1, \theta_2, \dots, \theta_n) \cdot \mathbf{p}_0 \right\},
\end{equation}

where \(\mathbf{p}_0\) is the origin point in the end-effector frame. We precompute these joint configurations offline, discretize the workspace into a voxel grid \(\mathcal{W}_{\text{voxel}}\), and store it for efficient computation in subsequent steps.

Next, as shown in Figure \ref{image2}, the robot's raw point cloud \(\mathcal{P}_{\text{raw}}\) is captured from its egocentric RGB-D camera in the camera coordinate system and transformed into the robot's coordinate system using the extrinsic calibration matrix \(\mathbf{E}\), resulting in the transformed point cloud \(\mathcal{P}\):

\begin{equation}
\mathcal{P} = \mathbf{E} \cdot \mathcal{P}_{\text{raw}}.
\end{equation}

To ensure physical feasibility, we perform a voxel grid lookup to determine whether each point in \(\mathcal{P}\) lies within the precomputed reachable workspace \(\mathcal{W}_{\text{voxel}}\):

\begin{equation}
\mathcal{P}_{\text{valid}} = \left\{ \mathbf{p} \in \mathcal{P} \ \bigg| \ \mathbf{p} \in \mathcal{W}_{\text{voxel}} \right\}.
\end{equation}

This step filters the point cloud to include only points within the robot's reachable workspace, abstracting the robot's physical reachability into a generalized spatial form.

Finally, we transform the valid point cloud \(\mathcal{P}_{\text{valid}}\) back into the camera coordinate system and use the camera's intrinsic parameters to project these points onto the image plane. We then mark the regions that comply with physical reachability on the original depth map. For areas that are not reachable, we apply a gray mask and outline their boundaries. The resulting S-P Map clearly highlights regions that are beyond the robot's physical reach, providing a unified and abstracted representation of reachability that is independent of the specific robot configuration. This allows the model to focus on the spatial constraints of the task, without needing to account for the detailed physical parameters of each robot.

\subsection{Model Architecture}
\label{3.2}

To seamlessly integrate robotic physical reachability into PhysVLM while preserving its visual reasoning capabilities, we design a dual-branch architecture: one branch dedicated to vision processing and the other to physical reachability (see Figure \ref{image2}). These branches operate independently, extracting features from their respective inputs, which are then fused and passed to a unified decoder for final reasoning and response generation.

The vision branch leverages a pre-trained Vision Transformer (ViT) \cite{vit}, specifically the \textit{SigLip-400M} model \cite{siglip}, to extract high-level visual features from egocentric images. To reduce computational overhead, a \textit{Max Pooling} layer is applied, followed by a two-layer Multi-Layer Perceptron (MLP) that transforms the visual features into token representations suitable for multimodal fusion.

The physical reachability branch processes the S-P Map, which abstracts the robot's physical reachability into a generalized spatial form. This branch also utilizes the \textit{SigLip-400M} model for feature extraction, followed by \textit{Max Pooling} and a feature fusion layer. The fusion layer combines the visual and reachability features, and a two-layer MLP further refines these fused features into reachability-specific tokens.

For the language decoding, we employ the \textit{Qwen-2.5-Instruct-3B} model \cite{qwen2, qwen2.5} as PhysVLM's large language model (LLM) decoder, using the \textit{Qwen-2.5} tokenizer to process natural language instructions. The decoder integrates multimodal tokens from the vision branch, S-P Maps, and language inputs, generating coherent and contextually relevant textual responses that account for both visual and physical reachability information.

\subsection{Training}
\label{3.3}

\begin{figure*}[ht] 
\vskip 0.2in
\begin{center}
\centerline{\includegraphics[width=1\linewidth]{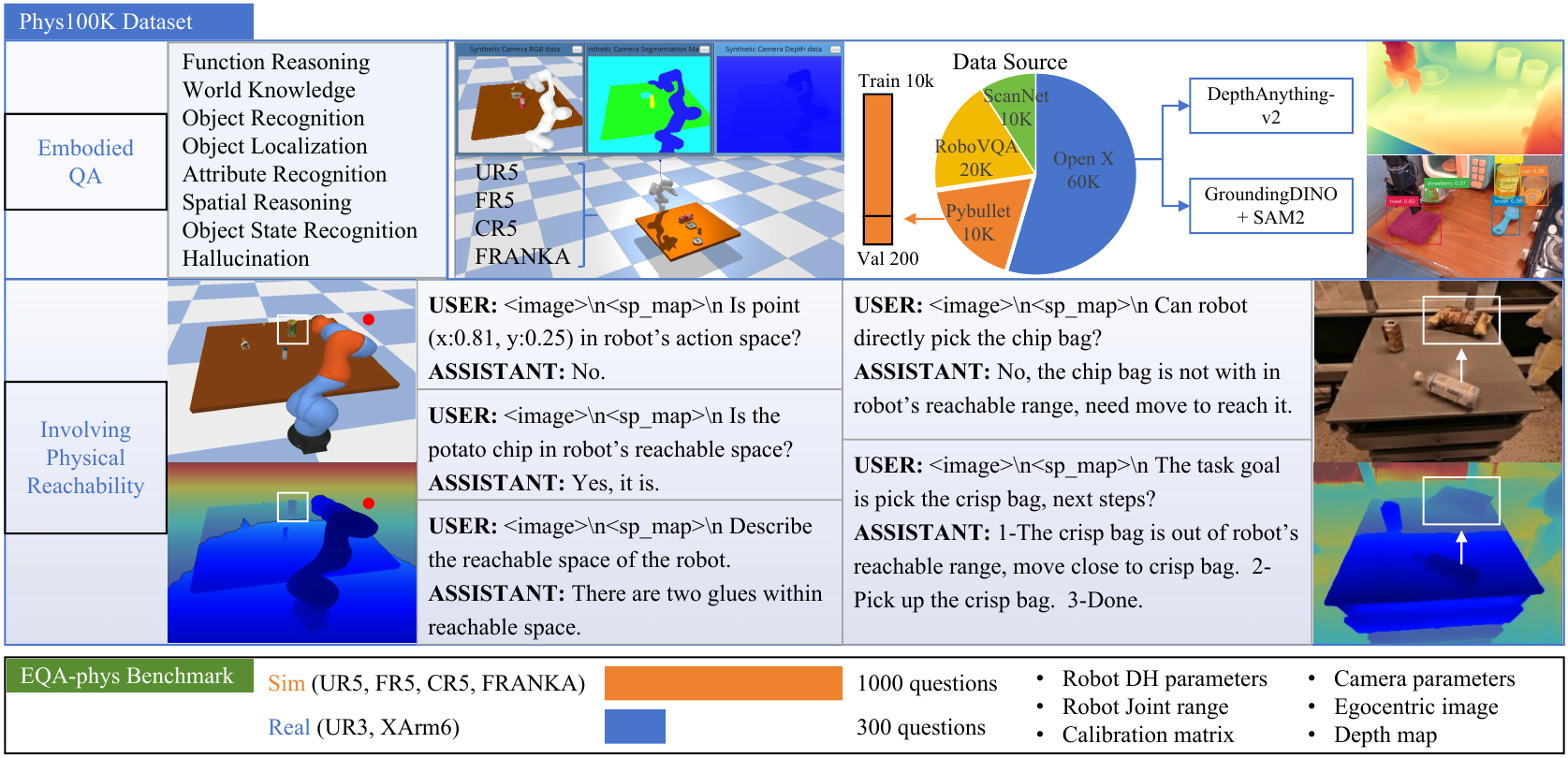}}
\caption{Details of the Phys100K Dataset and EQA-Phys Benchmark.}
\label{image5}
\end{center}
\vskip -0.2in
\end{figure*}

\paragraph{Training Data Construction.}
The training data for PhysVLM consists of our Phys100K dataset and general VQA datasets, such as LLaVA-Pretrain, ShareGPT4V, and RoboVQA. Phys100K focuses on question-answering related to physical reachability, aggregating data from RoboVQA (20K samples), ScanNet \cite{scannet} (10K samples), OpenX-Embodiment \cite{openx} (60K samples), and an additional 10K samples from PyBullet.

Depth maps are essential inputs for generating the S-P Map. For datasets lacking depth maps, we generate them using DepthAnything-v2. Additionally, we employ Grounding DINO \cite{groundingdino} and SAM2 \cite{sam2} to obtain 2D bounding boxes and segmentation results for objects in the images. In PyBullet, we simulate work scenarios using four robotic arms (UR5, FR5, CR5, and FRANKA) to collect RGB images, depth maps, and segmentation results.

For the PyBullet data in Phys100K, precise robot configurations can be obtained from the simulator. Therefore, we directly generate the S-P Map using the method described in Section \ref{3.2}, and labels indicating whether objects are reachable are obtained through simulated motion. The advantage of the S-P Map is that it abstracts physical reachability into a region-based representation, decoupling the learning process from specific robot configurations. This allows us to generate pseudo-labels for datasets without precise robot parameters. We approximate reachability using the segmentation results, marking regions and the objects within them as "reachable" or "unreachable" based on depth values. Next, we generate question-answer pairs for two main categories:

\begin{itemize}
\item \textbf{Embodied QA.} GPT-4 generates question-answer pairs for ScanNet and RoboVQA, covering categories include Function Reasoning, World Knowledge, Object Recognition, Object Localization, Attribute Recognition, Spatial Reasoning,  Object State Recognition, and Hallucination (See Figure \ref{image5}). Detailed prompts and examples are provided in the appendix to guide the generation process.
\item \textbf{Tasks Involving Physical Reachability.} We use the "reachable" label with five fixed task templates to generate question-answer pairs, such as \texttt{"\textbf{USER}:\textless image\textgreater\textbackslash n\textless sp\_map\textgreater\textbackslash n Is the [Object] in the robot's reachable space? \textbf{ASSISTANT}: Yes, it is."} Here, \texttt{[Object]} represents the relevant object category, while \texttt{\textless image\textgreater} and \texttt{\textless sp\_map\textgreater} serve as placeholders for image patch tokens and S-P Map patch tokens, respectively. Figure \ref{image5} provides examples of question-answer pairs for each object category.
\end{itemize}

\paragraph{Training Pipeline.}
We adopt a two-stage training process to fully leverage the S-P Map and ensure PhysVLM generalizes across different robots. In the first stage, we align multimodal features using the LLaVA-Pretrain and OpenX-Embodiment datasets from Phys100K. This stage only trains the projection layers, allowing the model to build a foundational understanding of visual inputs and physical reachability, independent of specific robot configurations.

In the second stage, we unfreeze all parameters and train the entire model using data from Phys100K, ShareGPT4V, and RoboVQA. This stage enhances PhysVLM's ability to handle complex visual reasoning tasks with physical reachability constraints, ensuring the model can generalize across diverse environments and robots.

\paragraph{Implementation Details.} 
PhysVLM is trained for 48 hours using eight A800 GPUs. The training process consists of two stages, each lasting one epoch. The batch size and learning rate are set at 128 and 1e-3 in the first stage, and 64 and 1e-5 in the second stage. The final model is PhysVLM-3B.

\subsection{EQA-phys Benchmark} 
As illustrated in Figure \ref{image5}, we introduce an embodied QA task focused on physical reachability, termed EQA-phys, which emphasizes QA tasks constrained by physical limitations. This benchmark includes a simulator dataset with 200 samples and 1,000 questions from the PyBullet validation set, as well as a zero-shot evaluation set based on real-world data from UR3 and XArm6 robots in two scenarios. The evaluation set contains 60 samples and 300 questions, all manually annotated by domain experts.

\begin{table*}[ht]
\caption{Results of EQA-phys. Comparison of PhysVLM-3B (ours) with API-based VLMs and embodied VLMs.}
\label{table1}
\begin{center}
\begin{small}
\begin{sc}
\begin{tabular}{llccccccc}
\toprule
 & & \multicolumn{2}{|c|}{Real-world} & \multicolumn{4}{|c|}{Simulator} & \\
 & & UR3 & XArm6 & UR5 & FR5 & CR5 & FRANKA & ALL \\
\midrule
\multirow{6}*{\makecell[l]{API-based\\VLMs}} & GPT-4o-mini & 54.3 & 56.0 & 49.4 & 55.4 & 54.6 & 47.1 & 52.8 \\
~ & Claude-3.5 & 56.2 & 60.5 & 54.0 & 58.1 & 55.7 & 54.3 & 56.4\\
~ & GPT-4o      & 56.7 & 61.5 & 55.7 & 58.3 & 57.5 & 52.6 & 57.0\\
~ & GPT-4o-mini \textbf{+ S-P Map} & $60.0_{\uparrow5.7}$ & $60.5_{\uparrow4.5}$ & $57.0_{\uparrow7.6}$ & $59.1_{\uparrow3.7}$ & $59.2_{\uparrow4.6}$ & $53.3_{\uparrow6.2}$ & $59.8_{\uparrow7.0}$\\
~ & Claude-3.5 \textbf{+ S-P Map} & $65.3_{\uparrow9.1}$ & $67.3_{\uparrow6.8}$ & $54.9_{\uparrow0.9}$ & $58.3_{\uparrow0.2}$ & $58.2_{\uparrow2.5}$ & $58.1_{\uparrow3.8}$ & $60.3_{\uparrow3.4}$\\
~ & GPT-4o \textbf{+ S-P Map}  & $\bm{66.6}_{\uparrow9.9}$ & $\bm{68.1}_{\uparrow6.6}$ & $55.8_{\uparrow0.1}$ & $60.7_{\uparrow1.4}$ & $59.4_{\uparrow1.9}$ & $57.6_{\uparrow5.0}$ & $61.3_{\uparrow4.1}$\\
\midrule
 \multirow{3}*{\makecell[l]{Embodied\\VLMs}} & SpatialVLM & 56.3 & 55.1 & 54.6 & 59.1 & 52.0 & 47.5 & 54.1 \\
~ & SpatialBot & 51.1 & 50.2 & 50.0 & 48.1 & 53.3 & 54.4 & 51.1 \\
~ & \textbf{PhysVLM-3B}  & 64.1 & 63.0 & \textbf{71.4} & \textbf{75.7} & \textbf{74.0} & \textbf{78.1} & \textbf{71.0} \\
\bottomrule
\end{tabular}
\end{sc}
\end{small}
\end{center}
\vskip -0.1in
\end{table*}

\section{Experiments}

\subsection{Experimental Setting}

\paragraph{Tasks.}
We compare the performance of PhysVLM with other methods across three categories of tasks:

\begin{itemize}
\item \textbf{EQA-phys.} This benchmark tests the model's ability to integrate visual reasoning with robotic physical reachability. The real-robot component is used to assess PhysVLM's zero-shot generalization, highlighting its ability to handle unseen robots and environments.
\item \textbf{Embodied QA.} We evaluate the model's general visual reasoning ability in embodied tasks using the OpenEQA \cite{openeqa} and RoboVQA-val \cite{robovqa} benchmarks.
\item \textbf{Robot Task Planning.} For real-world tasks such as \texttt{"Pick A into B"}, we assess the model's ability to understand robotic physical reachability and generate reasonable task plans. As this study does not focus on robot control strategies, we use the natural language planning approach from \cite{robovqa}.
\end{itemize}

\paragraph{Baselines.}
We compare our model to several baselines, including API-accessible VLMs such as Claude 3.5 \cite{claude3}, GPT-4o-mini \cite{gpt4}, and GPT-4o \cite{gpt4}, as well as embodied VLMs like SpatialVLM \cite{spatialvlm}, SpatialBot \cite{spatialbot}, 3D-VLA \cite{3dvla}, and RoboMamba \cite{robomamba}. SpatialVLM and SpatialBot both use the 3B version, which has a similar parameter count to our model. Since the executable versions of 3D-VLA and RoboMamba are unavailable, we compare their reported results on RoboVQA-val.

\paragraph{Evaluation Metrics.}
For tasks involving physical reachability, we use LLM scoring, following the approach used in existing studies \cite{openeqa, mmbench}. Assigning 5 points for completely correct responses and 1 point for incorrect responses, we calculate the average score and express it as a percentage. For Embodied QA, we follow the benchmark settings of the corresponding datasets \cite{openeqa, robovqa}. For task planning, each task type is executed 10 times, and the average success rate serves as the evaluation metric.

\subsection{Results on EQA-phys}
Table \ref{table1} shows the results on EQA-phys. Neither API-based nor embodied VLMs can handle the robot's parameter constraints, resulting in suboptimal outputs with scores around 55\%. In contrast, our model successfully completes visual reasoning tasks involving physical reachability, obtaining an average score of 71\%. As discussed, these tasks require the model to perform visual reasoning based on an understanding of the robotic physical reachability. The model can only effectively accomplish these tasks by truly understanding robotic physical reachability.

The results in Table \ref{table1} demonstrate that prompting API-based VLMs, such as GPT-4o, with the S-P Map (detailed in \ref{image6}) significantly enhances their performance. This improvement stems from the S-P Map's ability to abstract physical reachability into a robot-agnostic representation, enabling VLMs to reason about physical constraints that would otherwise be beyond their capabilities. By decoupling reachability from specific robot parameters, the S-P Map facilitates generalization across diverse environments, allowing models to better comprehend physical reachability, even in previously unseen scenarios.

Table \ref{table1} also demonstrates that PhysVLM-3B achieved scores of over 63\% in zero-shot evaluations for UR3 and XArm6 robots, despite operating in new environments with different robot parameters. This performance is attributed to two key factors: (1) the S-P Map abstracts various robot parameters into a unified, transferable representation of physical reachability, and (2) the model's independent visual and constraint encoding branches allow it to learn generalizable visual features from diverse image-text data, enabling effective reasoning in novel environments.

\begin{table}[ht]
\caption{Embodied QA results on the RoboVQA-val set, comparison of PhysVLM (ours) with existing methods. An asterisk (*) indicates models not pre-trained on the RoboVQA dataset.}
\label{table2}

\begin{center}
\begin{small}
\begin{sc}
\begin{tabular}{lcccc}
\toprule
 & BLEU1 & BLEU2 & BLEU3 & BLEU4 \\
\midrule
SpatialVLM* & 5.1 & 3.0 & 1.9 & 1.2  \\
SpatialBot* & 12.4 & 9.3 & 8.0 & 7.2  \\
3D-VLA     & 48.3  & 38.5  & 31.7  & 26.8   \\
RoboMamba  & 54.9  & 44.2  & 39.5  & 36.3   \\
\textbf{PhysVLM-3B} & \textbf{65.3}  & \textbf{62.4}  & \textbf{50.9}  & \textbf{43.5}   \\
\bottomrule
\end{tabular}
\end{sc}
\end{small}
\end{center}
\vskip -0.1in
\end{table}

\begin{table}[ht]
\caption{Embodied QA results on the OpenEQA benchmark, comparison of PhysVLM (ours) with existing methods. An asterisk (*) indicates that only the first 200 samples were tested due to API limitations.}
\label{table3}

\begin{center}
\begin{small}
\begin{sc}
\begin{tabular}{lccc}
\toprule
 & \makecell[l]{EM-EQA\\(ScanNet)} & \makecell[l]{EM-EQA\\(HM3D)} & ALL \\
\midrule
SpatialVLM & 42.9 & 44.3 & 43.8 \\
SpatialBot & 45.3 & 51.0 & 49.1 \\
GPT4V      & 57.4  & \underline{51.3}  & 55.3  \\
GPT-4o*     & \textbf{68.2}  & \textbf{65.2}  & \textbf{66.7} \\
\textbf{PhysVLM-3B} & \underline{60.7}  & 51.2  & \underline{57.4}  \\
\bottomrule
\end{tabular}
\end{sc}
\end{small}
\end{center}
\vskip -0.1in
\end{table}

\subsection{Results on Embodied QA}

We demonstrate our model's effectiveness in handling general embodied visual reasoning tasks. Additionally, we show that incorporating an understanding of physical constraints does not diminish its general visual reasoning capabilities. We compare our model with state-of-the-art embodied VLMs (see Tables \ref{table2} and \ref{table3}). Our model achieves the best performance on the RoboVQA-val benchmark, surpassing other models by 7.2\% in BLEU-4. On the OpenEQA benchmark, our model outperforms existing embodied VLMs and GPT-4V, ranking second, behind only GPT-4o.

\subsection{Results on Robot Task Planning}
Table \ref{table4} presents the performance of PhysVLM and baseline models on real-world task planning scenarios. When all objects are within the robot's physical reach, PhysVLM performs similarly to other models, as directly grabbing or placing the objects succeeds. However, when some objects are outside the physical reach, the model must suggest that the robot move closer before grabbing or placing them. In these cases, our model performs exceptionally well, whereas the task success rates of other models decline significantly. This is attributed to our model's understanding of robotic physical reachability and its ability to incorporate this understanding in task planning.

\begin{table}[ht]
\caption{Task planning results. Comparison of PhysVLM (ours) with other VLMs.}
\label{table4}

\begin{center}
\begin{small}
\begin{sc}
\begin{tabular}{lcc}
\toprule
 & \makecell[l]{All objects\\in range} & \makecell[l]{Part objects\\in range} \\
\midrule
GPT-4o-mini & 70.5   & 23.2  \\
Claude-3.5 & 73.6   & 32.1  \\
GPT-4o      & \textbf{75.9}   & 35.8  \\
SpatialVLM & 64.4   & 21.5  \\
SpatialBot & 65.6   & 25.3  \\
\textbf{PhysVLM-3B}    & 69.2   & \textbf{48.4} \\
\bottomrule
\end{tabular}
\end{sc}
\end{small}
\end{center}
\vskip -0.1in
\end{table}

\begin{table}[ht]
\caption{Ablation study on the S-P Map. We compare the constraint encoder's performance when using the S-P Map, replacing it with a Depth Map, or providing no input.}
\label{table5}

\begin{center}
\begin{small}
\begin{sc}
\begin{tabular}{lcccc}
\toprule
ID & \makecell[l]{S-P\\Map}   & \makecell[l]{Depth\\Map}  &  \makecell[l]{EQA-phys\\Real} & \makecell[l]{EQA-phys\\Sim}\\
\midrule
1 & \checkmark &  &  63.5 & 74.8 \\
2 & & \checkmark &  58.1 & 62.4\\
3 & & &  54.2 & 58.8 \\
\bottomrule
\end{tabular}
\end{sc}
\end{small}
\end{center}
\vskip -0.1in
\end{table}

\subsection{Ablation Study}
In this section, we conduct ablation studies to evaluate the contribution of each component in PhysVLM. We report the average LLM score on tasks involving physical reachability.

\paragraph{The effectiveness of S-P Map.} 
To demonstrate the S-P Map's contribution, we compare results obtained with and without its input. As shown in Table \ref{table5}, Experiments 1 and 3 demonstrate that omitting the S-P Map leads to a significant performance decrease for both zero-shot real-world robots and simulators. Specifically, the overall average score drops by 16\% in simulation results and 9.3\% in real-world robot evaluations. Without the S-P Map input, the model struggles to handle the robotic physical reachability.

Additionally, Experiments 1 and 2 demonstrate that replacing the S-P Map with a Depth Map significantly degrades the model's performance on zero-shot tasks. Since the Depth Map does not accurately represent robotic physical reachability, the model cannot rely solely on depth information to understand it.

\begin{table}[ht]
\caption{Ablation study on the effectiveness of an additional feature encoder. \texttt{Share} indicates shared network and weights with the visual feature encoder.}
\label{table6}

\begin{center}
\begin{small}
\begin{sc}
\begin{tabular}{lccc}
\toprule
    & EQA-phys & OpenEQA \\
\midrule
Independent & 71.0 & 57.4 & \\
Share       & 68.2 & 56.5 & \\
\bottomrule
\end{tabular}
\end{sc}
\end{small}
\end{center}
\vskip -0.1in
\end{table}

\begin{figure*}[ht] 
\vskip 0.2in
\begin{center}
\centerline{\includegraphics[width=0.94\linewidth]{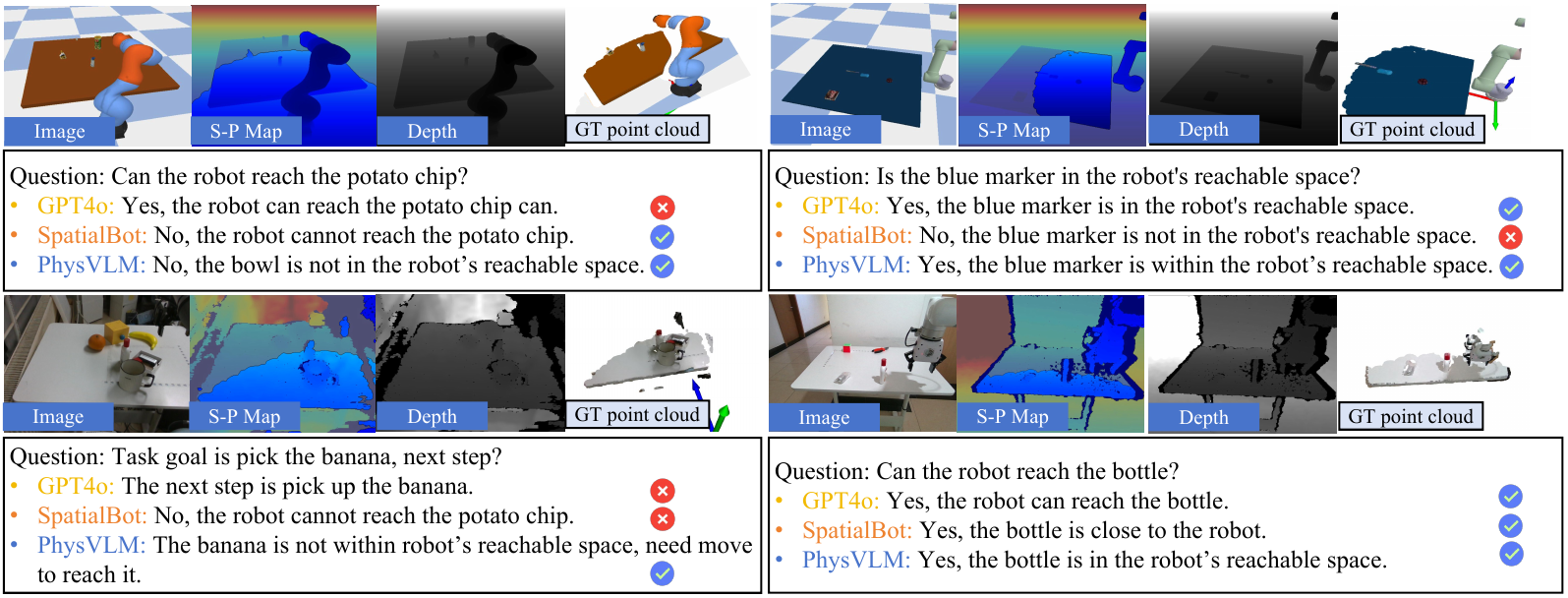}}
\caption{Visual comparison of PhysVLM (ours), GPT-4o, and SpatialBot.}
\label{image6}
\end{center}
\vskip -0.2in
\end{figure*}

\paragraph{Effectiveness of an additional feature encoder.}
To demonstrate the effectiveness of the model architecture, we compare the performance of a feature encoder that shares weights with the visual feature encoder for the S-P Map. In these experiments, we evaluate the average scores on both the EQA-phys and OpenEQA benchmarks. The results in Table \ref{table6} show that sharing the feature encoder not only decreases performance on EQA-phys but also impairs general visual reasoning capabilities. This is because S-P Map features differ from images, and the training data contains significantly more image-text pairs than S-P Map data.

\paragraph{Effectiveness of training data.}

To evaluate the effectiveness of Phys100K, we conduct experiments by selectively removing data from various sources in Phys100K. As shown in Table \ref{table7}, removing data from PyBullet or other embodied datasets leads to a reduction in overall performance, highlighting the critical role of each data component in the model's performance.

\begin{table}[ht]
\caption{Ablation study on the effectiveness of training data.}
\label{table7}

\begin{center}
\begin{small}
\begin{sc}
\begin{tabular}{lcc}
\toprule
Part of Phys100K & \makecell[l]{EQA-phys\\Real} & \makecell[l]{EQA-phys\\Sim} \\
\midrule
All & 63.5 & 74.8  \\
w/o Pybullet & 62.1 & 65.4  \\
w/o Other Datasets & 58.6 & 71.5  \\
\bottomrule
\end{tabular}
\end{sc}
\end{small}
\end{center}
\vskip -0.1in
\end{table}

\subsection{Qualitative Results}
Figure \ref{image6} compares our method with SpatialBot and GPT-4o. SpatialBot uses depth maps and images, while GPT-4o uses standard images. Both struggle with tasks requiring physical reachability, causing visual reasoning errors. In contrast, our approach delivers accurate results. Additionally, incorporating the S-P Map into GPT-4o improves its handling of physical reachability and response accuracy.

\section{Conclusion}
We introduce PhysVLM, a VLM that incorporates physical reachability into visual reasoning for robotic tasks. The S-P Map provides a unified representation of robotic reachability, facilitating the learning of generalizable features. PhysVLM extends traditional VLMs by adding a physical reachability encoder, enabling the simultaneous processing of visual, reachability, and textual information. Additionally, we present EQA-phys, a benchmark for evaluating embodied QA tasks involving physical reachability. Our experiments show that PhysVLM outperforms existing models, achieving a 14\% higher score than GPT-4o on EQA-phys. A limitation is its reduced zero-shot performance on real robots compared to simulations, likely due to the domain gap. Future work will focus on expanding datasets, enhancing real-world performance, and improving the understanding of physical accessibility in vision-language-action models. PhysVLM's reachability awareness supports safer and more reliable robotic decision-making in industrial and assistive settings, while its unified representation ensures cross-platform adaptability for real-world deployment, bridging crucial gaps between environmental perception and actionable robotic intelligence.

\section{Acknowledgments}
This work was supported by National Key R\&D Program of China under Grant No.2022ZD0160601, 2022YFB4300400, and 2018B030322016, National Natural Science Foundation of China under Grants 62176254, 62276260, and U1701266.

{
    \small
    \bibliographystyle{ieeenat_fullname}
    \bibliography{main}
}


\end{document}